# Analysis and Extension of Arc-Cosine Kernels for Large Margin Classification

Youngmin Cho*, Lawrence K. Saul

*Department of Computer Science and Engineering*
*University of California, San Diego*
*La Jolla, CA 92093, USA*

**Abstract**

We investigate a recently proposed family of positive-definite kernels that mimic the computation in large neural networks. We examine the properties of these kernels using tools from differential geometry; specifically, we analyze the geometry of surfaces in Hilbert space that are induced by these kernels. When this geometry is described by a Riemannian manifold, we derive results for the metric, curvature, and volume element. Interestingly, though, we find that the simplest kernel in this family does not admit such an interpretation. We explore two variations of these kernels that mimic computation in neural networks with different activation functions. We experiment with these new kernels on several data sets and highlight their general trends in performance for classification.

## 1. Introduction

Kernel methods provide a powerful framework for pattern analysis and classification (Schölkopf and Smola, 2001). The "kernel trick" works by mapping inputs into a nonlinear, potentially infinite-dimensional feature space, then applying classical linear methods in this space (Aizerman et al., 1964). The mapping is induced by a kernel function that operates on pairs of inputs and computes a generalized inner product. Typically, the kernel function measures some highly nonlinear or domain-specific notion of similarity.

*Corresponding author. Tel.:+1-858-699-2956.
   *Email addresses:* yoc002@cs.ucsd.edu (Youngmin Cho), saul@cs.ucsd.edu (Lawrence K. Saul)

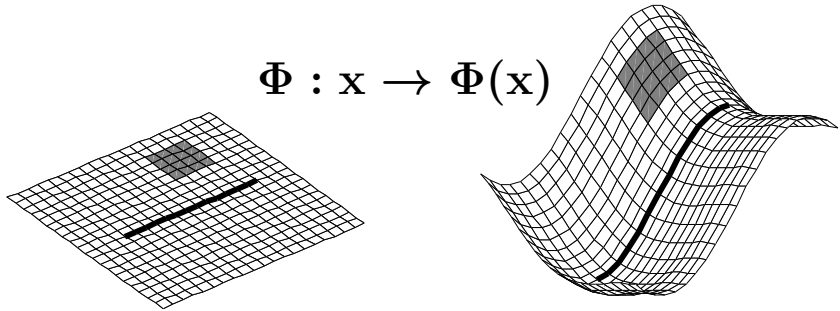

Figure 1: The kernel function induces a mapping from the input space into a nonlinear feature space. We can study the geometry of this surface in feature space—for example, asking how arc lengths and volume elements transform under this mapping.

Recently, Cho and Saul (2009, 2010) introduced a new family of positive-definite kernels that mimic the computation in large neural networks. These so-called "arc-cosine" kernels were derived by considering the mappings performed by infinitely large neural networks with Gaussian random weights and nonlinear threshold units (Williams, 1998). The kernels in this family can be viewed as computing inner products between inputs that have been transformed in this way.

Cho and Saul (2009, 2010) experimented with these kernels on various benchmark data sets for deep learning (Larochelle et al., 2007). On some data sets, these kernels surpassed the best previous results from deep belief nets, suggesting that many advantages of neural networks might ultimately be incorporated into kernel-based methods (Weston et al., 2008). Such an intriguing possibility seems worth exploring given the respective advantages of these competing approaches to machine learning (Bengio and LeCun, 2007; Bengio, 2009).

In this paper, we investigate the geometric properties of arc-cosine kernels in much greater detail. Specifically, we analyze the geometry of surfaces in Hilbert space that are induced by these kernels. These surfaces are the images of the input space under the implicit nonlinear mapping performed by the kernel; see Fig. 1. Our analysis yields a richer understanding of the geometry of these surfaces (and by association, the nonlinear transformations parameterized by large neural networks). We also compare and contrast our results to those previously obtained for Gaussian and polynomial kernels (Amari and Wu, 1999; Burges, 1999).

As one important theoretical contribution, our analysis shows that arc-cosine kernels of different degrees have qualitatively different geometric properties. In particular, for some kernels in this family, the surface in Hilbert space is described by a curved Riemannian manifold; for another kernel, this surface is flat, with zero intrinsic curvature; finally, for the simplest member of the family, this surface cannot be described as a manifold at all. It seems that the family of arc-cosine kernels exhibits a larger variety of behaviors than other popular families of kernels.

Our work also explores new, related kernels that are derived from large neural networks with different activation functions. The original arc-cosine kernels were derived by considering the mappings in neural networks with Heaviside step functions. We derive two new kernels by examining the effects of either shifting or smoothing the discontinuities of these step functions. The first of these operations (biasing) induces more sparse representations of the data in feature space, while the second (smoothing) removes the non-analyticity of the simplest kernel in the arc-cosine family. Both effects are interesting to explore given the improvements they have yielded in conventional neural networks. We evaluate these variations of arc-cosine kernels in support vector machines for large margin classification (Boser et al., 1992; Cortes and Vapnik, 1995; Cristianini and Shawe-Taylor, 2000). Our experiments show that these variations of arc-cosine kernels often lead to better performance.

This paper is organized as follows. In section 2, we analyze the surfaces in Hilbert spaces induced by arc-cosine kernels and derive expressions for the metric, volume element, and scalar curvature when these surfaces can be described as Riemannian manifolds. In section 3, we show how to construct new kernels by considering neural networks with biased or smoothed activation functions. In section 4, we present our experimental results. Finally, in section 5, we summarize our most important findings and suggest various directions for future research.

## 2. Analysis

In this section, we review the family of positive-definite kernels introduced by Cho and Saul (2009, 2010) and examine their properties using tools from differential geometry (Amari and Wu, 1999; Burges, 1999). Specifically, we analyze the geometry of surfaces in Hilbert space that are induced by these kernels. When this geometry is described by a Riemannian manifold, we derive results for the metric, curvature, and volume element. We also examine a kernel in this family that does not admit such an interpretation.

## 2.1. Arc-cosine kernels

We briefly review the basic form of *arc-cosine* kernels. The $n$th-order kernel in this family is defined by the integral representation

$$k_n(\mathbf{x}, \mathbf{y}) = 2\int d\mathbf{w}\, \frac{e^{-\frac{\|\mathbf{w}\|^2}{2}}}{(2\pi)^{d/2}} \Theta(\mathbf{w}\cdot\mathbf{x})\Theta(\mathbf{w}\cdot\mathbf{y})(\mathbf{w}\cdot\mathbf{x})^n(\mathbf{w}\cdot\mathbf{y})^n \qquad (1)$$

where $\Theta(z) = \frac{1}{2}[1+\text{sign}(z)]$ denotes the Heaviside step function and $n$ is restricted to be a nonnegative integer. Interestingly, the kernel function $k_n(\mathbf{x}, \mathbf{y})$ in eq. (1) mimics the computation in a large neural network with Gaussian random weights and nonlinear threshold units. In particular, it can be viewed as computing the inner product between the images of the inputs $\mathbf{x}$ and $\mathbf{y}$ after they have been transformed by a network of this form. The particular form of the threshold nonlinearity is determined by the value of $n$.

The integral in eq. (1) can be done analytically. In particular, let $\theta$ denote the angle between the inputs $\mathbf{x}$ and $\mathbf{y}$:

$$\theta = \cos^{-1}\left(\frac{\mathbf{x}\cdot\mathbf{y}}{\|\mathbf{x}\|\|\mathbf{y}\|}\right). \qquad (2)$$

For the case $n=0$, the kernel function in eq. (1) takes the simple form:

$$k_0(\mathbf{x}, \mathbf{y}) = 1 - \frac{\theta}{\pi}. \qquad (3)$$

For the general case, the $n$th order kernel function in this family can be written as:

$$k_n(\mathbf{x}, \mathbf{y}) = \frac{1}{\pi}\|\mathbf{x}\|^n\|\mathbf{y}\|^n J_n(\theta), \qquad (4)$$

where all the angular dependence is captured by the functions $J_n(\theta)$. These functions are given by:

$$J_n(\theta) = (-1)^n(\sin\theta)^{2n+1}\left(\frac{1}{\sin\theta}\frac{\partial}{\partial\theta}\right)^n\left(\frac{\pi-\theta}{\sin\theta}\right). \qquad (5)$$

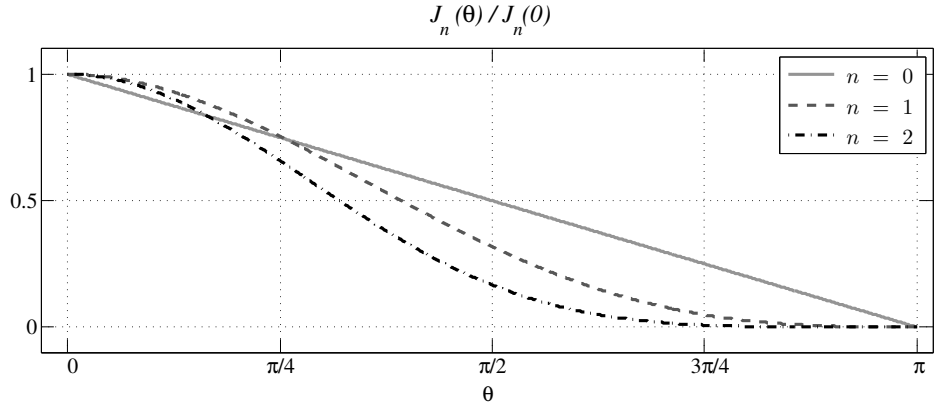

Figure 2: Behavior of the function $J_n(\theta)$ in eq. (5) for small values of $n$.

The so-called *arc-cosine* kernels in eq. (4) are named for their nontrivial dependence on the angle $\theta$ and the arc-cosine function in eq. (2).

Fig. 2 plots the functions $J_n(\theta)$ for small values of $n$. The first three expressions of $J_n(\theta)$ are:

$$J_0(\theta) = \pi - \theta, \tag{6}$$
$$J_1(\theta) = \sin\theta + (\pi - \theta)\cos\theta, \tag{7}$$
$$J_2(\theta) = 3\sin\theta\cos\theta + (\pi - \theta)(1 + 2\cos^2\theta). \tag{8}$$

Note how these expressions exhibit a different dependence on the angle $\theta$ than a purely linear kernel, which can be written as $k(\mathbf{x}, \mathbf{y}) = \|\mathbf{x}\|\|\mathbf{y}\|\cos\theta$. In general, the function $J_n(\theta)$ takes its maximum value at $\theta=0$ and decreases monotonically to zero at $\theta=\pi$. However, as shown in the figure, the derivative $J'_n(\theta)$ at $\theta=0$ only vanishes for positive integers $n \geq 1$.

## 2.2. Riemannian geometry

We can understand the family of arc-cosine kernels better by analyzing the geometry of surfaces in Hilbert space. For surfaces that can be described as Riemannian manifolds, Burges (1999) and Amari and Wu (1999) showed how to derive the metric, volume element, and curvature directly from the kernel function. In this section, we use these methods to study arc-cosine kernels of degree $n \geq 1$. As some of the calculations are lengthy, we sketch the main results here while providing more detailed derivations in the appendix.

*2.2.1. Metric*

We briefly review the relation of the metric to the kernel function $k(\mathbf{x}, \mathbf{y})$. Consider the surface in Hilbert space parameterized by the input coordinates $x_\mu$. The line element on the surface is given by:

$$ds^2 = \|\mathbf{\Phi}(\mathbf{x}+\mathbf{dx}) - \mathbf{\Phi}(\mathbf{x})\|^2$$
$$= k(\mathbf{x}+\mathbf{dx}, \mathbf{x}+\mathbf{dx}) - 2k(\mathbf{x}, \mathbf{x}+\mathbf{dx}) + k(\mathbf{x}, \mathbf{x}). \tag{9}$$

We identify the metric $g_{\mu\nu}$ by expanding the right hand side to second order in the displacement $\mathbf{dx}$. In terms of the metric, the line element is given by:

$$ds^2 = g_{\mu\nu} dx^\mu dx^\nu, \tag{10}$$

where a sum over repeated indices is implied. Finally, equating the last two expression gives:

$$g_{\mu\nu} = \frac{1}{2} \frac{\partial}{\partial x_\mu} \frac{\partial}{\partial x_\nu} k(\mathbf{x}, \mathbf{x}) - \left[ \frac{\partial}{\partial y_\mu} \frac{\partial}{\partial y_\nu} k(\mathbf{x}, \mathbf{y}) \right]_{\mathbf{y}=\mathbf{x}} \tag{11}$$

provided that the kernel function $k(\mathbf{x}, \mathbf{y})$ is twice-differentiable.

We now consider the metrics induced by the family of arc-cosine kernels $k_n(\mathbf{x}, \mathbf{y})$ in eq. (4) of degree $n \geq 1$. As a first step, we analyze the behavior of these kernels for nearby inputs $\mathbf{x} \approx \mathbf{y}$. This behavior is in turn determined by the behavior of the functions $J_n(\theta)$ in eq. (5) for small values of $\theta$. For $n \geq 1$, this behavior is locally quadratic with a maximum at $\theta = 0$. In particular, by expanding the integral representation in eq. (1) for small values of $\theta$, it can be shown that:

$$J_n(0) = \pi (2n-1)!!, \tag{12}$$
$$J_n(\theta) \approx J_n(0) \left(1 - \frac{n^2 \theta^2}{2(2n-1)}\right), \tag{13}$$

where $(2n-1)!! = \frac{(2n)!}{2^n n!}$ is known as the double-factorial function. Together with eq. (4), the quadratic expansion in eq. (13) captures the behavior of the arc-cosine kernels $k_n(\mathbf{x}, \mathbf{y})$ for nearby inputs $\mathbf{x} \approx \mathbf{y}$. It follows from eq. (11) that this behavior also determines the form of the metric.

The metrics for arc-cosine kernels of degree $n \geq 1$ can be derived by substituting the general form in eq. (4) into eq. (11). After some algebra (see appendix), this calculation gives:

$$g_{\mu\nu} = n^2 (2n-3)!! \|\mathbf{x}\|^{2n-2} \left( \delta_{\mu\nu} + 2(n-1) \frac{x_\mu x_\nu}{\|\mathbf{x}\|^2} \right). \tag{14}$$

Table 1: Comparison of the metric $g_{\mu\nu}$ and scalar curvature $S$ for different kernels over inputs $\mathbf{x} \in \Re^d$. The results for polynomial (the second row) and Gaussian kernels (the third row) were derived by Burges (1999). The results for arc-cosine kernels (the bottom row) are valid for kernels of order $n \geq 1$.

| $k(\mathbf{x}, \mathbf{y})$ | $g_{\mu\nu}$ | $S$ |
|---|---|---|
| $\mathbf{x} \cdot \mathbf{y}$ | $\delta_{\mu\nu}$ | $0$ |
| $(\mathbf{x} \cdot \mathbf{y})^p$ | $p \, \|\mathbf{x}\|^{2p-2} \left( \delta_{\mu\nu} + (p{-}1) \frac{x_\mu x_\nu}{\|\mathbf{x}\|^2} \right)$ | $\frac{(p{-}1)(2{-}d)(d{-}1)}{p\|\mathbf{x}\|^{2p}}$ |
| $e^{-\|\mathbf{x}-\mathbf{y}\|^2/\sigma^2}$ | $(2/\sigma^2)\delta_{\mu\nu}$ | $0$ |
| $\frac{\|\mathbf{x}\|^n \|\mathbf{y}\|^n}{\pi} J_n(\theta)$ | $n^2(2n{-}3)!! \, \|\mathbf{x}\|^{2n-2} \left( \delta_{\mu\nu} + 2(n{-}1)\frac{x_\mu x_\nu}{\|\mathbf{x}\|^2} \right)$ | $\frac{3(n{-}1)^2 \, (2{-}d) \, (d{-}1)}{n^2 \, (2n{-}1)!! \, \|\mathbf{x}\|^{2n}}$ |

In general, this metric differs from the Euclidean metric in the prefactor $\|\mathbf{x}\|^{2n-2}$ and the projection component $x_\mu x_\nu / \|\mathbf{x}\|^2$. However, it is worth distinguishing two qualitatively different regimes of behavior.

*Case $n = 1$: the manifold is flat.*

Though the metric in eq. (14) is generally non-Euclidean, an interesting result emerges in the special case $n = 1$: it reduces to the Euclidean metric $g_{\mu\nu} = \delta_{\mu\nu}$. Thus, despite performing a highly nonlinear mapping, the $n = 1$ arc-cosine kernel induces a surface in Hilbert space that is intrinsically flat. Previously, we observed that the $n = 1$ arc-cosine kernel preserves the norm of inputs (Cho and Saul, 2009, 2010), with $k_1(\mathbf{x}, \mathbf{x}) = \|\mathbf{x}\|^2$. In this paper, we have shown that also like the purely linear kernel, it preserves the fully Euclidean metric.

In fact, it is not unusual for nonlinear kernels to preserve the Euclidean metric. Burges (1999) observed that all translationally invariant kernels of the form $k(\mathbf{x}, \mathbf{y}) = k(\mathbf{x}-\mathbf{y})$ have this property, including the popular family of Gaussian kernels $k(\mathbf{x}, \mathbf{y}) = e^{-\|\mathbf{x}-\mathbf{y}\|^2/\sigma^2}$. Note, however, that the $n=1$ arc-cosine kernel is not translationally invariant. It represents a different form of nonlinearity that nonetheless preserves the Euclidean metric.

*Case $n \geq 2$: the manifold is curved.*

The metrics from arc-cosine kernels of degree $n \geq 2$ have a similar form as metrics from homogeneous polynomial kernels of degree $p \geq 2$. Table 1 compares our results to previous results obtained for polynomial and Gaus-sian kernels (Amari and Wu, 1999; Burges, 1999). We will see later that the metric in eq. (14) describes a manifold with non-zero intrinsic curvature if the inputs $\mathbf{x} \in \Re^d$ live in three or more dimensions ($d > 2$).

*2.2.2. Volume element*

The metric $g_{\mu\nu}$ determines other interesting quantities as well. For example, the volume element $dV$ on the manifold is given by:

$$dV = \sqrt{\det g_{\mu\nu}}\, d\mathbf{x}. \tag{15}$$

Assuming that the mapping from inputs to features is one-to-one, the volume element determines how a probability density transforms under this mapping.

The determinant of the metric for arc-cosine kernels is straightforward to compute. In particular, noting that the metric in eq. (14) is proportional to the identity matrix plus a projection matrix, we find:

$$\det(g) = (2n-1)\left(n^2(2n-3)!!\,\|\mathbf{x}\|^{2n-2}\right)^d. \tag{16}$$

For the special case $n = 1$, this expression reduces to $\det(g) = 1$, consistent with the previous observation that in this case, the metric is Euclidean.

*2.2.3. Curvature*

The metric $g_{\mu\nu}$ also determines the intrinsic curvature of the manifold. The curvature is expressed in terms of the Christoffel elements of the second kind:

$$\Gamma^{\alpha}_{\beta\gamma} = \frac{1}{2}g^{\alpha\mu}\left(\partial_\beta g_{\gamma\mu} - \partial_\mu g_{\beta\gamma} + \partial_\gamma g_{\mu\beta}\right), \tag{17}$$

where $\partial_\mu = \partial/\partial x_\mu$ denotes the partial derivative and $g^{\alpha\mu}$ denotes the matrix inverse of the metric. In terms of these quantities, the Riemann curvature tensor is given by:

$$R_{\nu\alpha\beta}{}^{\mu} = \partial_\alpha \Gamma^{\mu}_{\beta\nu} - \partial_\beta \Gamma^{\mu}_{\alpha\nu} + \Gamma^{\rho}_{\alpha\nu}\Gamma^{\mu}_{\beta\rho} - \Gamma^{\rho}_{\beta\nu}\Gamma^{\mu}_{\alpha\rho}. \tag{18}$$

The elements of $R_{\nu\alpha\beta}{}^{\mu}$ vanish for a manifold with no intrinsic curvature. The scalar curvature is given by:

$$S = g^{\nu\beta} R_{\nu\mu\beta}{}^{\mu}. \tag{19}$$

The scalar curvature describes the amount by which the volume of a geodesic ball on the manifold deviates from that of a ball in Euclidean space.

Substituting the metric in eq. (14) into eqs. (17–19), we obtain the scalar curvature for surfaces in Hilbert space induced by arc-cosine kernels:

$$S = \frac{3(n-1)^2\,(2-d)\,(d-1)}{n^2\,(2n-1)!!\,\|\mathbf{x}\|^{2n}}. \tag{20}$$

Note that the curvature vanishes for the kernel of degree $n=1$, as well as for all kernels in this family when the inputs $\mathbf{x} \in \Re^d$ lie in $\Re^1$ or $\Re^2$. The vanishing of the curvature in these circumstances is also observed in the family of homogeneous polynomial kernels, as shown in Table 1.

2.3. Non-analytic kernels

Next we show that the $n=0$ arc-cosine kernel does not induce a surface in Hilbert space whose geometry can be described as a Riemannian manifold. Consider the squared distance between the feature vectors $\Phi(\mathbf{x})$ and $\Phi(\mathbf{x}+\mathbf{dx})$ induced by this kernel for an infinitesimal displacement $\mathbf{dx}$. To begin, we compute this distance for non-radial displacements $\mathbf{dx}_\perp$ that are orthogonal to $\mathbf{x}$, satisfying $\mathbf{x} \cdot \mathbf{dx}_\perp = 0$. Recall that the $n=0$ arc-cosine kernel maps all inputs $\mathbf{x}$ to the unit hypersphere in feature space, with $k_0(\mathbf{x}, \mathbf{x}) = 1$. Exploiting this property, we find:

$$\begin{aligned}
\|\Phi(\mathbf{x}+\mathbf{dx}_\perp) &- \Phi(\mathbf{x})\|^2 \\
&= k_0(\mathbf{x}+\mathbf{dx}_\perp, \mathbf{x}+\mathbf{dx}_\perp) + k_0(\mathbf{x},\mathbf{x}) - 2k_0(\mathbf{x}, \mathbf{x}+\mathbf{dx}_\perp) \\
&= (2/\pi)\cos^{-1}(\|\mathbf{x}\|/\|\mathbf{x}+\mathbf{dx}_\perp\|) \\
&= (2/\pi)\sin^{-1}(\|\mathbf{dx}_\perp\|/\|\mathbf{x}+\mathbf{dx}_\perp\|) \\
&\approx (2/\pi)\|\mathbf{dx}_\perp\|/\|\mathbf{x}\|,
\end{aligned} \tag{21}$$

where in the last line we have approximated the right hand side by its first-order Taylor series and kept only leading terms.

To generalize eq. (21) to arbitrary displacements, we note that the $n=0$ arc-cosine kernel is invariant to the magnitude of its arguments, depending only on the angle $\theta$ between them. Thus, eq. (21) generalizes easily to displacements $\mathbf{dx}$ that include a radial component. In particular, we can write:

$$\|\Phi(\mathbf{x}+\mathbf{dx}) - \Phi(\mathbf{x})\|^2 = \frac{2}{\pi\|\mathbf{x}\|}\sqrt{\mathbf{dx}^\top\left(\mathbf{I}_d - \frac{\mathbf{x}\mathbf{x}^\top}{\|\mathbf{x}\|^2}\right)\mathbf{dx}}, \tag{22}$$

which merely projects out the radial component before computing the infinitesimal squared distance; here $\mathbf{I}_d$ is the $d\times d$ identity matrix.

Note that the right hand side of eq. (22) does not have the form of a Riemannian metric. In particular, the infinitesimal squared distance in feature space scales *linearly* not quadratically with $\|\mathbf{dx}_\perp\|$. This behavior arises from the non-analyticity of the arc-cosine function, which does not admit a Taylor series expansion around its root at unity: $\cos^{-1}(1-\epsilon) \approx \sqrt{2\epsilon}$ for $0 < \epsilon \ll 1$. This non-analyticity not only distinguishes the $n=0$ arc-cosine kernel from higher-order kernels in this family, but also from all polynomial and Gaussian kernels.

### 3. Extensions

In this section, we explore two variations on the arc-cosine kernel of degree $n=0$. Specifically, we construct new kernels by modifying the Heaviside step functions $\Theta(\cdot)$ that appear in eq. (1). We consider the effects of shifting the thresholds of these step functions as well as smoothing their nonlinearities. These modifications introduce new parameters—measuring the amount of shift or smoothing—that can be tuned to improve the performance of the resulting kernels.

*3.1. Biased threshold functions*

Consider the arc-cosine kernel of degree $n = 0$ as defined by eq. (1). We obtain a new kernel by translating the Heaviside step functions in this definition by a bias term $b \in \Re$:

$$k^b(\mathbf{x}, \mathbf{y}) = 2 \int \mathbf{dw}\, \frac{e^{-\frac{\|\mathbf{w}\|^2}{2}}}{(2\pi)^{d/2}}\, \Theta(\mathbf{w} \cdot \mathbf{x} - b)\, \Theta(\mathbf{w} \cdot \mathbf{y} - b). \qquad (23)$$

The motivation behind this construction is to regulate the sparsity of the infinite dimensional representation $\mathbf{\Phi}(\mathbf{x})$. Note that as the bias $b$ is increased, a larger volume of the weight space $\mathbf{w} \in \Re^d$ is associated with zero activation levels $\Theta(\mathbf{w} \cdot \mathbf{x} - b)$ from the input $\mathbf{x} \in \Re^d$. Thus this construction is able to emulate a large neural network with especially sparse ($b>0$) or dense ($b<0$) hidden unit representations.

The integral in eq. (23) cannot be performed in closed form, but we can express it in terms of simple one-dimensional definite integrals. To this end, we use $\xi$, $\psi$, and $\theta$ to denote the internal angles of the triangle formed by the vectors $\mathbf{x}$, $\mathbf{y}$, and $\mathbf{x} - \mathbf{y}$; see Fig. 3. Also, as shorthand, we define the

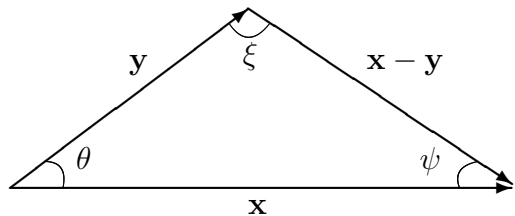

Figure 3: A triangle formed by the input data vectors **x**, **y**, and their difference **x** − **y**.

two-parameter family of definite integrals:

$$I(r, \xi) = \frac{1}{\pi} \int_0^\xi d\phi \, \exp\left(-\frac{1}{2\, r^2 \sin^2 \phi}\right). \quad (24)$$

It is simple to compute this integral and store the results in a lookup table for discretized values of $\xi \in [0, \pi]$ and $r > 0$.

We begin by evaluating the right hand side of eq. (23) in the regime $b \geq 0$ of increased sparsity. Then, in terms of the above notation, we obtain the result:

$$k^b(\mathbf{x}, \mathbf{y}) = I\left(b^{-1}\|\mathbf{x}\|, \psi\right) + I(b^{-1}\|\mathbf{y}\|, \xi) \quad \text{for} \quad b \geq 0. \quad (25)$$

The derivation of this result is given in the appendix.

The result in the opposite regime $b \leq 0$ is obtained by a simple transformation. In this regime, we can evaluate the integral in eq. (23) by noting that $\Theta(z) = 1 - \Theta(-z)$ and exploiting the symmetry of the integral in weight space. It follows that:

$$k^b(\mathbf{x}, \mathbf{y}) = k^{-b}(\mathbf{x}, \mathbf{y}) + \mathrm{erf}\left(\frac{-b}{\sqrt{2}\|\mathbf{x}\|}\right) + \mathrm{erf}\left(\frac{-b}{\sqrt{2}\|\mathbf{y}\|}\right) \quad \text{for} \quad b \leq 0, \quad (26)$$

where $\mathrm{erf}(x) = \frac{2}{\sqrt{\pi}} \int_0^x e^{-t^2} dt$ is the error function. From the same observations, it also follows that kernel matrices for opposite values of $b$ are equivalent up to centering (i.e., after subtracting out the mean in feature space). Thus without loss of generality, we only investigate kernels with biases $b \geq 0$ in our experiments on support vector machines (Boser et al., 1992; Cortes and Vapnik, 1995).

As already noted, the arc-cosine kernel of degree $n = 0$ depends only on the angle between its inputs and not on their magnitudes. The kernel in eq. (25) does not exhibit this same invariance. However, it does have the scaling property:

$$k^b(\rho\mathbf{x}, \rho\mathbf{y}) = k^{b/\rho}(\mathbf{x}, \mathbf{y}) \quad \text{for} \quad \rho > 0. \tag{27}$$

Eq. (27) shows that the effect of a different bias can be mimicked by uniformly rescaling all the inputs.

*3.2. Smoothed threshold functions*

We can extend the arc-cosine kernel of degree $n=0$ in a different way by smoothing the Heaviside step function in eq. (1). The simplest smooth alternative is the cumulative Gaussian function:

$$\Psi_\sigma(z) = \frac{1}{\sqrt{2\pi\sigma^2}} \int_{-\infty}^{z} du\, e^{-\frac{u^2}{2\sigma^2}}, \tag{28}$$

which reduces to the Heaviside step function in the limit of vanishing variance ($\sigma^2 \to 0$). The resulting kernel is defined as:

$$k_\sigma(\mathbf{x}, \mathbf{y}) = 2\int d\mathbf{w}\, \frac{e^{-\frac{\|\mathbf{w}\|^2}{2}}}{(2\pi)^{d/2}}\, \Psi_\sigma(\mathbf{w}\cdot\mathbf{x})\, \Psi_\sigma(\mathbf{w}\cdot\mathbf{y}) \tag{29}$$

The variance parameter $\sigma^2$ can be tuned in this kernel just as its counterpart in a radial basis function (RBF) kernel. However, note that RBF kernels behave very differently than these kernels in the limit of vanishing variance: the former become degenerate, whereas eq. (29) reduces to the arc-cosine kernel of degree $n=0$.

The integral in eq. (29) can be performed analytically, yielding the result:

$$k_\sigma(\mathbf{x}, \mathbf{y}) = 1 - \frac{1}{\pi} \cos^{-1}\left(\frac{\mathbf{x}\cdot\mathbf{y}}{\sqrt{(\|\mathbf{x}\|^2 + \sigma^2)(\|\mathbf{y}\|^2 + \sigma^2)}}\right). \tag{30}$$

Details of the calculation are given in the appendix. The kernel in eq. (30) is analogous to one derived earlier by Williams (1998) in the context of Gaussian processes. However, in that work, the kernel was computed for an activation function bounded between -1 and 1 (as opposed to 0 and 1, above).

One effect of smoothing the threshold function in eq. (29) is to remove the non-analyticity of the arc-cosine kernel of degree $n=0$, as described in section 2.3. It is straightforward to compute the Riemannian metric and volume element associated with the kernel in eq. (30):

$$g_{\mu\nu} = \frac{1}{\pi\sigma\sqrt{2\|\mathbf{x}\|^2 + \sigma^2}} \left( \delta_{\mu\nu} - \frac{2\mathbf{x}_\mu \mathbf{x}_\nu}{2\|\mathbf{x}\|^2 + \sigma^2} \right), \quad (31)$$

$$\det(g) = \frac{1}{\pi^d \sigma^{d-2} (2\|\mathbf{x}\|^2 + \sigma^2)^{\frac{d}{2}+1}}. \quad (32)$$

Table 2: Data set specifications: the number of training, validation, and test examples, input dimensionality, and the number of classes.

| Data set | Training | Validation | Test | Dimension | Class |
|---|---|---|---|---|---|
| MNIST-rand | 10000 | 2000 | 50000 | 784 | 10 |
| MNIST-image | 10000 | 2000 | 50000 | 784 | 10 |
| Rectangles | 1000 | 200 | 50000 | 784 | 2 |
| Rectangles-image | 10000 | 2000 | 50000 | 784 | 2 |
| Convex | 6000 | 2000 | 50000 | 784 | 2 |
| 20-Newsgroups | 12748 | 3187 | 3993 | 62061 | 20 |
| ISOLET | 4990 | 1248 | 1559 | 617 | 26 |
| Gisette | 4800 | 1200 | 1000 | 5000 | 2 |

Two observations are worth making here. First, from eq. (31), we see that the metric diverges as $\sigma$ vanishes, reflecting the non-analyticity of the arc-cosine kernel of degree $n=0$. Second, from eq. (32), we see that the volume element shrinks with increasing distance from the origin in input space (i.e., with increasing $\|\mathbf{x}\|$); this property distinguishes this kernel from all the other kernels in section 2.2.

## 4. Experimental results

We evaluated the new kernels in section 3 by measuring their performance in support vector machines (SVMs). We also compared them to other popular kernels for large margin classification.

*4.1. Data sets*

Table 2 lists the eight data sets used in our experiments. The top five data sets in the table are image classification benchmarks from an empirical evaluation of deep learning (Larochelle et al., 2007). The first two of these are noisy variations of the MNIST data set of handwritten digits (LeCun and Cortes, 1998): the task in *MNIST-rand* is to recognize digits whose backgrounds have been corrupted by white noise, while the task in *MNIST-image* is to recognize digits whose backgrounds consist of other image patches. The other benchmarks are purely synthetic data sets. The task in *Rectangles* is to classify a single rectangle that appears in each image as tall or wide. *Rectangles-image* is a harder variation of this task in which the background of each rectangle consists of other image patches. Finally, the task in *Convex* is to classify a single white region that appears in each image as convex or non-convex. We partitioned these data sets into training, validation, and test examples as in previous benchmarks.

The bottom three data sets in Table 2 are from benchmark problems in text categorization, spoken letter recognition, and feature selection. The task in *20-Newsgroups* is to classify newsgroup postings (represented as bags of words) into one of twenty news categories (Lang, 1995). The task in *ISOLET* is to identify a spoken letter of the English alphabet (Frank and Asuncion, 2010). The task in *Gisette* is to distinguish the MNIST digits FOUR versus NINE, but the input representation has been padded with a large number of additional features—some helpful, some spurious, and some sparse (Guyon et al., 2005). We randomly held out 20% of the training examples in these data sets for validation.

*4.2. Methodology*

For classification by SVMs, we compared five different kernels—two without tuning parameters and three with tuning parameters. Those without tuning parameters included the linear kernel and the arc-cosine kernel of degree $n=0$. Those with tuning parameters included the radial basis function (RBF) kernel, parameterized by its kernel width, as well as the variations on arc-cosine kernels in sections 3.1 and 3.2, parameterized by either the bias $b$ or variance $\sigma^2$.

All SVMs were trained using libSVM (Chang and Lin, 2001), a publicly available software package. For multiclass problems, we adopted the so-called one-versus-one approach: SVMs were trained for each pair of different classes, and test examples were labeled by the majority vote of all the pairwise SVMs.

We followed the same experimental methodology as in previous work (Larochelle et al., 2007; Cho and Saul, 2009) to tune the margin-violation penalties in SVMs as well as the kernel parameters. We used the held-out (validation) examples to determine these values, first searching over a coarse logarithmic grid, then performing a fine-grained search to improve their settings. Once these values were determined, however, we retrained each SVM on the combined set of training and validation examples. We used these retrained SVMs for the final performance evaluations on test examples.

Table 3: Classification error rates (%) on test sets from SVMs with various kernels. The first three kernels are the arc-cosine kernel of degree $n\!=\!0$ and the variations on this kernel described in sections 3.1 and 3.2. The best performing kernel for each data set is marked in bold. When different, the best performing arc-cosine kernel is marked in italics. See text for details.

| Data set | Arc-cosine | | | RBF | Linear |
| --- | --- | --- | --- | --- | --- |
| | $n\!=\!0$ | Bias | Smooth | | |
| MNIST-rand | 17.16 | *16.49* | 17.03 | **14.80** | 17.31 |
| MNIST-image | 23.81 | *23.77* | 24.09 | **22.80** | 25.07 |
| Rectangles | 13.08 | *2.48* | 11.84 | **2.11** | 30.30 |
| Rectangles-image | **22.66** | 23.59 | 24.48 | 23.42 | 49.69 |
| Convex | 20.05 | 20.12 | *19.60* | **18.76** | 45.67 |
| 20-Newsgroups | 16.28 | 16.25 | **15.73** | 15.75 | 15.90 |
| ISOLET | 3.40 | *3.34* | 3.53 | **3.01** | 3.53 |
| Gisette | **1.80** | 1.90 | 1.90 | 2.10 | 2.20 |

*4.3. Results*

Table 3 displays the test error rates from the experiments. In the majority of cases, the parameterized variations of arc-cosine kernels achieve better performance than the original arc-cosine kernel of degree $n = 0$. The gains demonstrate the utility of the variations based on biased or smoothed threshold functions. Most often, however, it remains true that the best results are still obtained from RBF kernels.

In previous work, we showed that the performance of arc-cosine kernels could be improved by a recursive composition (Cho and Saul, 2009, 2010) that mimicked the computation in multilayer neural networks. We experimented with the same procedure here using the variations of arc-cosine kernels described in sections 3.1 and 3.2. In these experiments, we deployed the arc-cosine kernels in Table 3 at the first layer of nonlinearity[1] and the arc-cosine kernel of degree $n\!=\!1$ at five subsequent layers of nonlinearity. The resulting error rates are shown in Table 4. In general, the composition of arc-cosine kernels again leads to improved performance, although RBF kernels still obtain the best performance on half of the data sets. The table reveals an interesting trend: we observe the improvements from composition mainly on the data sets that are not sparse (such as *20-Newsgroups* and *Gisette*). It seems that sparse data sets do not lend themselves as well to the construction of multilayer kernels.

Table 4: Classification error rates (%) on the test set for arc-cosine kernels and their multilayer extensions. The best performing kernel for each data set is marked in bold. When different, the best performing arc-cosine kernel is marked in italics. See text for details.

| Data set | Arc-cosine | | | Arc-cosine multilayer | | | RBF |
|---|---|---|---|---|---|---|---|
| | $n\!=\!0$ | Bias | Smth | $n\!=\!0$ | Bias | Smth | |
| MNIST-rand | 17.16 | 16.49 | 17.03 | 16.21 | *16.1* | 16.96 | **14.8** |
| MNIST-image | 23.81 | 23.77 | 24.09 | 23.15 | *23.1* | 23.3 | **22.8** |
| Rectangles | 13.08 | *2.48* | 11.84 | 6.76 | 2.88 | 5.57 | **2.11** |
| Rectangles-image | 22.66 | 23.59 | 24.48 | **22.35** | 22.6 | 23.18 | 23.42 |
| Convex | 20.05 | 20.12 | 19.6 | 19.09 | 18.79 | **18.29** | 18.76 |
| 20-Newsgroups | 16.28 | 16.25 | **15.73** | 16.8 | 17.13 | 15.93 | 15.75 |
| ISOLET | 3.4 | 3.34 | 3.53 | 3.34 | 3.27 | *3.14* | **3.01** |
| Gisette | **1.8** | 1.9 | 1.9 | 1.9 | 2.1 | **1.8** | 2.1 |

## 5. Discussion

In this paper, we have investigated the geometric properties of arc-cosine kernels and explored variations of these kernels with additional tuning parameters. The geometric properties were studied by analyzing the surfaces that these kernels induced in Hilbert space. Here, we observed the following: (i) for arc-cosine kernels of degree $n \geq 2$, these surfaces are curved Rieman-nian manifolds (like those from polynomial kernels of degree $p \geq 2$); (ii) for the arc-cosine kernel of degree $n=1$, this surface has vanishing scalar curvature (like those from linear and Gaussian kernels); and (iii) for the arc-cosine kernel of degree $n=0$, this surface cannot be described as a Riemannian manifold due to the non-analyticity of the kernel function. Our main theoretical contributions are summarized in Table 1.

---

[1] We used the same bias and smoothness parameters that were determined previously for the experiments of Table 3.

We also explored variations of arc-cosine kernels that were designed to mimic the computations in large neural networks with biased or smoothed activation functions. We evaluated these new kernels extensively for large margin classification in SVMs. By tuning the bias and variance parameters in these kernels, we showed that they often performed better than the original arc-cosine kernel of degree $n=0$. Many of these results were further improved when these new kernels were composed with other arc-cosine kernels to mimic the computations in multilayer neural networks. On some data sets, these multilayer kernels yielded lower error rates than the best performing RBF kernels.

Our theoretical and experimental results suggest many possible directions for future work. One direction is to leverage the geometric properties of the arc-cosine kernels for better classification performance. Such an idea was proposed earlier by Amari and Wu (1999) and Wu and Amari (2002), who used a conformal transformation to increase the spatial resolution around the decision boundary induced by RBF kernels. The volume elements in eq. (16) and eq. (32) allow us to explore similar methods for the kernels analyzed in this paper.

Given the relatively simple form of the volume element, another possible direction is to explore the use of arc-cosine kernels for probabilistic modeling. Such an approach might exploit the connection with neural computation to provide a kernel-based alternative to inference and learning in deep belief networks (Hinton et al., 2006). Though our experimental results have not revealed a clear connection between the geometric properties of arc-cosine kernels and their performance in SVMs, it is worth emphasizing that kernels are used in many settings beyond classification, including clustering, dimensionality reduction, and manifold learning. In these other settings, the geometric properties of arc-cosine kernels may play a more prominent role.

Finally, we are interested in more effective schemes to combine and compose arc-cosine kernels. Additive combinations of kernels have been studied in the framework of multiple kernel learning (Lanckriet et al., 2004). For multiple kernel learning with arc-cosine kernels, the base kernels could vary in the degree $n$ as well as the bias $b$ and variance $\sigma^2$ parameters introduced in section 3. Composition of these kernels should also be fully explored as this operation (applied repeatedly) can be used to mimic the computations in different multilayer neural nets. We are studying these issues and others in ongoing work.

## Acknowledgements

This work was supported by award number 0957560 from the National Science Foundation. The authors also thank Fei Sha for suggesting to consider the kernel in section 3.1.

## Appendix  A. Derivation of Riemannian metric

In this appendix, we show how to derive the results for the Riemannian metric and curvature that appear in section 2.2. We begin by deriving the individual terms that appear in the expression for the metric in eq. (11). Substituting the form of the arc-cosine kernels in eq. (4) into this expression, we obtain:

$$\partial_{x_\mu}\partial_{x_\nu}\left[k_n(\mathbf{x},\mathbf{x})\right] = \frac{2}{\pi}\|\mathbf{x}\|^{2n-2} J_n(0)\left[n\,\delta_{\mu\nu} + 2n(n-1)\frac{x_\mu x_\nu}{\|\mathbf{x}\|^2}\right] \quad (A.1)$$

$$\partial_{y_\mu}\partial_{y_\nu}\left[k_n(\mathbf{x},\mathbf{y})\right]_{\mathbf{y}=\mathbf{x}} = \frac{1}{\pi}\|\mathbf{x}\|^{2n-2}\left(J_n(0)\left[n\,\delta_{\mu\nu} + n(n-2)\frac{x_\mu x_\nu}{\|\mathbf{x}\|^2}\right]\right.$$
$$\left. + \left[\frac{J_n'(\theta)}{\sin\theta}\right]_{\theta=0}\left[\delta_{\mu\nu} - \frac{x_\mu x_\nu}{\|\mathbf{x}\|^2}\right]\right) \quad (A.2)$$

where $\partial_{x_\mu}$ is shorthand for the partial derivative with respect to $x_\mu$. To complete the derivation of the metric, we must evaluate the terms $J_n(0)$ and $\lim_{\theta\to 0}\left[J_n'(\theta)/\sin\theta\right]$ that appear in these expressions. As shown in previous work (Cho and Saul, 2009), an expression for $J_n(\theta)$ is given by the two-dimensional integral:

$$J_n(\theta) = \int_{-\infty}^{\infty} dw_1 \int_{-\infty}^{\infty} dw_2 \left[e^{-\frac{w_1^2+w_2^2}{2}}\,\Theta(w_1)\,\Theta(w_1\cos\theta + w_2\sin\theta)\right.$$
$$\left. \times w_1^n (w_1\cos\theta + w_2\sin\theta)^n\right]. \quad (A.3)$$

It is straightforward to evaluate this integral at $\theta = 0$, which yields the result in eq. (12). Differentiating under the integral sign and evaluating at $\theta = 0$, we obtain:

$$J'_n(0) = n \int_{-\infty}^{\infty} dw_1 \int_{-\infty}^{\infty} dw_2 \ e^{-\frac{w_1^2+w_2^2}{2}} \ \Theta(w_1) \ w_1^{2n-1} \ w_2 = 0, \quad (A.4)$$

where the integral vanishes due to symmetry. To evaluate the rightmost term in eq. (A.2), we avail ourselves of l'Hôpital's rule:

$$\lim_{\theta \to 0} \left[ \frac{J'_n(\theta)}{\sin \theta} \right] = \lim_{\theta \to 0} J''_n(\theta) = J''_n(0). \quad (A.5)$$

Differentiating eq. (A.3) twice under the integral sign and evaluating at $\theta = 0$, we obtain:

$$\begin{aligned} J''_n(0) &= n \int_{-\infty}^{\infty} dw_1 \int_{-\infty}^{\infty} dw_2 \ e^{-\frac{w_1^2+w_2^2}{2}} \ \Theta(w_1) \ w_1^{2n-2} \left[(n-1)w_2^2 - w_1^2\right] \\ &= -\pi n^2 (2n-3)!!. \end{aligned} \quad (A.6)$$

Substituting these results into eq. (11), we obtain the expression for the metric in eq. (14). The remaining calculations for the curvature are tedious but straightforward. Using the Woodbury matrix identity, we can compute the matrix inverse of the metric as:

$$g^{\mu\nu} = \frac{1}{\|\mathbf{x}\|^{2n-2} n^2 (2n-3)!!} \left( \delta_{\mu\nu} - \frac{x_\mu x_\nu}{\|\mathbf{x}\|^2} \frac{2(n-1)}{2n-1} \right). \quad (A.7)$$

The partial derivatives of the metric are also easily computed as:

$$\begin{aligned} \partial_\beta g_{\gamma\mu} &= 2n^2(n-1)(2n-3)!! \ \|\mathbf{x}\|^{2n-4} \\ &\quad \times \left( x_\beta \delta_{\gamma\mu} + x_\gamma \delta_{\beta\mu} + x_\mu \delta_{\beta\gamma} + (2n-4)\frac{x_\beta x_\gamma x_\mu}{\|\mathbf{x}\|^2} \right). \end{aligned} \quad (A.8)$$

Substituting these results for the metric inverse and partial derivatives into eq. (17), we obtain the Christoffel elements of the second kind:

$$\Gamma^\alpha_{\beta\gamma} = \frac{n-1}{\|\mathbf{x}\|^2} \left( x_\beta \delta_{\alpha\gamma} + x_\gamma \delta_{\alpha\beta} + \frac{x_\alpha \delta_{\beta\gamma}}{2n-1} - \frac{2n}{2n-1} \frac{x_\alpha x_\beta x_\gamma}{\|\mathbf{x}\|^2} \right). \quad (A.9)$$

Substituting these Christoffel elements into eq. (18), we obtain the Riemann curvature tensor:

$$R_{\nu\alpha\beta}{}^{\mu} = \frac{3}{\|\mathbf{x}\|^2} \frac{(n-1)^2}{2n-1} \left( \frac{x_\mu x_\alpha \delta_{\beta\nu} - x_\mu x_\beta \delta_{\nu\alpha} + x_\nu x_\beta \delta_{\mu\alpha} - x_\nu x_\alpha \delta_{\mu\beta}}{\|\mathbf{x}\|^2} \right. \\ \left. + \delta_{\mu\beta}\delta_{\nu\alpha} - \delta_{\mu\alpha}\delta_{\beta\nu} \right). \quad (A.10)$$

Finally, combining eqs. (A.7) and (A.10), we obtain the scalar curvature $S$ in eq. (20).

## Appendix B. Derivation of kernel with biased threshold functions

In this appendix we show how to evaluate the integral in eq. (23). As in previous work (Cho and Saul, 2009), we start by adopting coordinates in which $\mathbf{x}$ aligns with the $w_1$ axis and $\mathbf{y}$ lies in the $w_1 w_2$-plane:

$$\mathbf{x} = \mathbf{e_1} \|\mathbf{x}\|, \quad (B.1)$$
$$\mathbf{y} = (\mathbf{e_1} \cos\theta + \mathbf{e_2} \sin\theta) \|\mathbf{y}\|, \quad (B.2)$$

where $\mathbf{e_i}$ is the unit vector along the $i$th axis and $\theta$ is defined in eq. (2). Next we substitute these expressions into eq. (23) and integrate out the remaining orthogonal coordinates of the weight vector $\mathbf{w}$. What remains is the two dimensional integral:

$$k^b(\mathbf{x}, \mathbf{y}) = \frac{1}{\pi} \int_{-\infty}^{\infty} dw_1 \int_{-\infty}^{\infty} dw_2 \left[ e^{-\frac{w_1^2 + w_2^2}{2}} \right. \\ \left. \times \Theta(w_1 \|\mathbf{x}\| - b) \, \Theta(w_1 \|\mathbf{y}\| \cos\theta + w_2 \|\mathbf{y}\| \sin\theta - b) \right]. \quad (B.3)$$

We can simplify this further by adopting polar coordinates $(r, \phi)$ in the $w_1 w_2$-plane of integration, where $w_1 = r \cos\phi$ and $w_2 = r \sin\phi$. With this change of variables, we obtain the polar integral:

$$k^b(\mathbf{x}, \mathbf{y}) = \frac{1}{\pi} \int_{-\pi}^{\pi} d\phi \int_0^{\infty} dr \left[ r e^{-\frac{r^2}{2}} \right. \\ \left. \times \Theta(r\|\mathbf{x}\| \cos\phi - b) \, \Theta(r\|\mathbf{y}\| \cos(\theta - \phi) - b) \right]. \quad (B.4)$$

This integral can be evaluated in the feasible region of the plane that is defined by the arguments of the step functions. In what follows, we assume $b > 0$ since the opposite case can be derived by symmetry (as shown in section 3.1). We identify the feasible region by conditioning the arguments of the step functions to be positive:

$$\cos\phi \;>\; 0, \tag{B.5}$$

$$\cos(\phi - \theta) \;>\; 0, \tag{B.6}$$

$$r \;>\; \max\left(\frac{b}{\|\mathbf{x}\|\cos\phi}, \frac{b}{\|\mathbf{y}\|\cos(\phi-\theta)}\right). \tag{B.7}$$

The first two of these inequalities limit the range of the angular integral; in particular, we require $\theta - \frac{\pi}{2} < \phi < \frac{\pi}{2}$. The third bounds the range of the radial integral from below. We can perform the radial integral analytically to obtain:

$$k^b(\mathbf{x},\mathbf{y}) \;=\; \frac{1}{\pi}\int_{\theta-\frac{\pi}{2}}^{\frac{\pi}{2}} d\phi \,\min\left[e^{-\frac{1}{2}\left(\frac{b}{\|\mathbf{x}\|\cos\phi}\right)^2},\; e^{-\frac{1}{2}\left(\frac{b}{\|\mathbf{y}\|\cos(\phi-\theta)}\right)^2}\right] \tag{B.8}$$

The term that is selected by the minimum operation in eq. (B.8) depends on the value of $\phi$. The crossover point $\phi_c$ occurs where the exponents are equal, namely at $\|\mathbf{x}\|\cos\phi_c = \|\mathbf{y}\|\cos(\phi_c - \theta)$. Solving for $\phi_c$ yields:

$$\phi_c \;=\; \tan^{-1}\left(\frac{\|\mathbf{x}\|}{\|\mathbf{y}\|\sin\theta} - \cot\theta\right). \tag{B.9}$$

To disentangle the min-operation in the integrand, we break the range of integration into two intervals:

$$k^b(\mathbf{x},\mathbf{y}) \;=\; \frac{1}{\pi}\int_{\theta-\frac{\pi}{2}}^{\phi_c} d\phi\, e^{-\frac{1}{2}\left(\frac{b}{\|\mathbf{y}\|\cos(\phi-\theta)}\right)^2} + \frac{1}{\pi}\int_{\phi_c}^{\frac{\pi}{2}} d\phi\, e^{-\frac{1}{2}\left(\frac{b}{\|\mathbf{x}\|\cos\phi}\right)^2} \tag{B.10}$$

Finally, we obtain a more symmetric expression by appealing to the angles $\xi$ and $\psi$ defined in Fig. 3. (Note that $\phi_c$ and $\psi$ are complementary angles, with $\phi_c = \frac{\pi}{2} - \psi$.) Writing eq. (B.10) in terms of the angles $\xi$ and $\psi$ yields the final form in eq. (25).

# Appendix C. Derivation of kernel with smoothed threshold functions

A simple transformation of the integral in eq. (29) reduces it to essentially the same integral as eq. (1). We begin by appealing to the integral representation of the cumulative Gaussian function:

$$\Psi_\sigma(\mathbf{w} \cdot \mathbf{x}) = \frac{1}{\sqrt{2\pi\sigma^2}} \int_{-\infty}^{\infty} d\mu \ e^{-\frac{\mu^2}{2\sigma^2}} \ \Theta(\mathbf{w} \cdot \mathbf{x} - \mu), \tag{C.1}$$

$$\Psi_\sigma(\mathbf{w} \cdot \mathbf{y}) = \frac{1}{\sqrt{2\pi\sigma^2}} \int_{-\infty}^{\infty} d\nu \ e^{-\frac{\nu^2}{2\sigma^2}} \ \Theta(\mathbf{w} \cdot \mathbf{y} - \nu). \tag{C.2}$$

After substituting these representations into eq. (29), we obtain an expanded integral over the weight vector $\mathbf{w}$ and the new auxiliary variables $\mu$ and $\nu$. Let $\bar{\mathbf{w}} \in \Re^{d+2}$ denote the extended weight vector obtained by appending two new elements to $\mathbf{w} \in \Re^d$ as follows:

$$\bar{\mathbf{w}} = \left(\mathbf{w}, \ \mu\sigma^{-1}, \nu\sigma^{-1}\right). \tag{C.3}$$

Also let $\bar{\mathbf{x}} \in \Re^{d+2}$ and $\bar{\mathbf{y}} \in \Re^{d+2}$ denote the extended inputs defined by appending two new elements to $\mathbf{x} \in \Re^d$ and $\mathbf{y} \in \Re^d$ as follows:

$$\bar{\mathbf{x}} = (\mathbf{x}, \ -\sigma, \ 0), \tag{C.4}$$
$$\bar{\mathbf{y}} = (\mathbf{y}, \ 0, \ -\sigma). \tag{C.5}$$

The transformation is completed by writing the required integral for eq. (29) in terms of $\bar{\mathbf{w}}, \bar{\mathbf{x}},$ and $\bar{\mathbf{y}}$. This change of variables yields an integral analogous to eq. (1), with $\bar{\mathbf{w}}, \bar{\mathbf{x}},$ and $\bar{\mathbf{y}}$ playing the same roles as $\mathbf{w}, \mathbf{x},$ and $\mathbf{y}$. The result in eq. (30) follows.

## References

Aizerman, M. A., Braverman, E. M., & Rozonoér, L. I. (1964). Theoretical foundations of the potential function method in pattern recognition learning. *Automation and Remote Control*, *25*:821–837.

Amari, S. & Wu, S. (1999). Improving support vector machine classifiers by modifying kernel functions. *Neural Networks*, *12*(6):783–789.

Bengio, Y. (2009). Learning deep architectures for AI. *Foundations and Trends in Machine Learning*, *2*(1):1–127.

Bengio, Y. & LeCun, Y. (2007). Scaling learning algorithms towards AI. In Bottou, L., Chapelle, O., Decoste, D., & Weston, J., editors, *Large-Scale Kernel Machines*. MIT Press.

Boser, B. E., Guyon, I. M., & Vapnik, V. N. (1992). A training algorithm for optimal margin classifiers. In *Proceedings of the Fifth Annual ACM Workshop on Computational Learning Theory*, pages 144–152. ACM Press.

Burges, C. J. C. (1999). Geometry and invariance in kernel based methods. In Schölkopf, B., Burges, C. J. C., & Smola, A., editors, *Advances in Kernel Methods - Support Vector Learning*. MIT Press, Cambridge.

Chang, C.-C. & Lin, C.-J. (2001). *LIBSVM: a library for support vector machines*. Software available at http://www.csie.ntu.edu.tw/~cjlin/libsvm.

Cho, Y. & Saul, L. K. (2009). Kernel methods for deep learning. In Bengio, Y., Schuurmans, D., Lafferty, J., Williams, C., & Culotta, A., editors, *Advances in Neural Information Processing Systems 22*, pages 342–350, Cambridge, MA. MIT Press.

Cho, Y. & Saul, L. K. (2010). Large-margin classification in infinite neural networks. *Neural Computation*, *22*(10):2678–2697.

Cortes, C. & Vapnik, V. (1995). Support-vector networks. *Machine Learning*, *20*:273–297.

Cristianini, N. & Shawe-Taylor, J. (2000). *An Introduction to Support Vector Machines and Other Kernel-based Learning Methods*. Cambridge University Press.

Frank, A. & Asuncion, A. (2010). UCI machine learning repository.

Guyon, I., Gunn, S., Ben-Hur, A., & Dror, G. (2005). Result analysis of the nips 2003 feature selection challenge. In Saul, L. K., Weiss, Y., & Bottou, L., editors, *Advances in Neural Information Processing Systems 17*, pages 545–552, Cambridge, MA. MIT Press.

Hinton, G. E., Osindero, S., & Teh, Y. W. (2006). A fast learning algorithm for deep belief nets. *Neural Computation*, *18*(7):1527–1554.

Lanckriet, G., Cristianini, N., Bartlett, P., Ghaoui, L. E., & Jordan, M. I. (2004). Learning the kernel matrix with semidefinite programming. *Journal of Machine Learning Research*, *5*:27–72.

Lang, K. (1995). Newsweeder: Learning to filter netnews. In *Proceedings of the 12th International Conference on Machine Learning (ICML-95)*, pages 331–339. Morgan Kaufmann publishers Inc.: San Mateo, CA, USA.

Larochelle, H., Erhan, D., Courville, A., Bergstra, J., & Bengio, Y. (2007). An empirical evaluation of deep architectures on problems with many factors of variation. In *Proceedings of the 24th International Conference on Machine Learning (ICML-07)*, pages 473–480.

LeCun, Y. & Cortes, C. (1998). The MNIST database of handwritten digits. http://yann.lecun.com/exdb/mnist/.

Schölkopf, B. & Smola, A. J. (2001). *Learning with Kernels: Support Vector Machines, Regularization, Optimization, and Beyond*. MIT Press, Cambridge, MA.

Weston, J., Ratle, F., & Collobert, R. (2008). Deep learning via semi-supervised embedding. In *Proceedings of the 25th International Conference on Machine Learning (ICML-08)*, pages 1168–1175.

Williams, C. K. I. (1998). Computation with infinite neural networks. *Neural Computation, 10(5)*:1203–1216.

Wu, S. & Amari, S. (2002). Conformal transformation of kernel functions: a data-dependent way to improve support vector machine classifiers. *Neural Processing Letters, 15(1)*:59–67.